\documentclass[preprint,12pt]{elsarticle}

\usepackage{amsmath,amssymb,amsfonts}
\usepackage{graphicx}
\usepackage{cleveref}
\usepackage{booktabs}
\usepackage{algpseudocode}
\usepackage{algorithm}
\usepackage[table]{xcolor}
\usepackage{balance}

\definecolor{mygreen}{RGB}{0, 150, 0}

\newcommand{\mypar}[1]{\noindent\textbf{#1.}}

\journal{Computerized Medical Imaging and Graphics}

\begin{document}

\begin{frontmatter}



\title{Temporally-Aware Diffusion Model for Brain Progression Modelling with Bidirectional Temporal Regularisation}

\affiliation[label1]{organization={University of Catania},
             city={Catania},
             country={Italy}}
\affiliation[label2]{organization={University of Nottingham},
             city={Nottingham},
             country={United Kingdom}}

\affiliation[label3]{organization={University of Messina},
             city={Messina},
             country={Italy}}

\author[label1]{Mattia Litrico} 
\author[label1]{Francesco Guarnera} 
\author[label2]{Mario Valerio Giuffrida} 
\author[label3]{Daniele Ravì} 
\author[label1]{Sebastiano Battiato} 

\begin{abstract}
Generating realistic MRIs to accurately predict future changes in the structure of brain is an invaluable tool for clinicians in assessing clinical outcomes and analysing the disease progression at the patient level. However, current existing methods present some limitations: (i) some approaches fail to explicitly capture the relationship between structural changes and time intervals, especially when trained on age-imbalanced datasets; (ii) others rely only on scan interpolation, which lack clinical utility, as they generate intermediate images between timepoints rather than future pathological progression; and (iii) most approaches rely on 2D slice-based architectures, thereby disregarding full 3D anatomical context, which is essential for accurate longitudinal predictions. We propose a 3D Temporally-Aware Diffusion Model (TADM-3D), which accurately predicts brain progression on MRI volumes. To better model the relationship between time interval and brain changes, TADM-3D uses a pre-trained \textit{Brain-Age Estimator} (BAE) that guides the diffusion model in the generation of MRIs that accurately reflect the expected age difference between baseline and generated follow-up scans. Additionally, to further improve the temporal awareness of TADM-3D, we propose the \textit{Back-In-Time Regularisation} (BITR), by training TADM-3D to predict bidirectionally from the baseline to follow-up (forward), as well as from the follow-up to baseline (backward). Although predicting past scans has limited clinical applications, this regularisation helps the model generate temporally more accurate scans.
We train and evaluate TADM-3D on the OASIS-3 dataset, and we validate the generalisation performance on an external test set from the NACC dataset. The code will be available upon acceptance.
\end{abstract}




\end{frontmatter}



\section{Introduction}
\label{sec:introduction}
Predicting how brain structures evolve over time in MRI scans is a key challenge in medical imaging. The modelling of temporal brain trajectories has proven useful in multiple applications, including recovering missing scans in longitudinal data \cite{ravi2019degenerative}, acting as a virtual placebo \cite{ravi2022degenerative}, aiding in patient stratification \cite{counterfactual_pred,ravi2019degenerative,ravi2022degenerative,Young2024}, and supporting both diagnosis and prognosis of Alzheimer’s disease (AD) \cite{bowles2018modelling}. Although AD diagnosis is often based on neuropsychological and behavioural evaluations, imaging data plays a crucial role in revealing structural brain changes associated with the disease, even at early stages \cite{jack2018nia}.

Recent advancements in AI have led to the emergence of novel spatio-temporal technologies designed to model disease progression \cite{liu2006spatial}, enabling more accurate predictions of structural brain changes. Among these, generative approaches have gained attention for their ability to accurately simulate future MRI scans based on earlier scans. Early methods used Generative Adversarial Networks (GANs) to generate future MRIs \cite{counterfactual_pred,ravi2022degenerative}. Recently, Denoising Diffusion Probabilistic Models (DDPMs) have become the state-of-the-art for generative models, demonstrating superior performance also in this domain. Sequence-Aware Diffusion Model (SADM) \cite{SADM} combines a diffusion model with a transformer to predict follow-up MRIs. Similarly, Diffusion Deformable Model (DDM) \cite{DDM} and DiffuseMorph \cite{DiffuseMorph} use diffusion models to estimate the deformation field between scans. Differently, BrLP \cite{lemuel24} proposes the training of a latent diffusion model in conjunction with an autoencoder and an auxiliary model to generate individual MRI scans based on longitudinal data.

However, these methods often show poor performance in accurately modelling the temporal dynamics of morphological changes under neurodegenerative diseases. Some approaches~\cite{counterfactual_pred,ravi2022degenerative,xia2019consistent,xia2020learning} do not explicitly capture the relationship between structural MRI changes and time intervals, while others are limited in their ability to generate future scans~\cite{DDM,DiffuseMorph} or require longitudinal data at inference time~\cite{SADM}.

To address these challenges, we introduce TADM-3D, a novel 3D diffusion-based framework specifically designed to predict future brain MRI scans. Our method learns the statistical distribution of brain changes over specified time intervals, capturing the complex, non-linear patterns of neuroanatomical trajectories directly on MRI data. Rather than directly generating follow-up MRI scans, which can be prone to artefacts, TADM-3D focuses on predicting the voxel-wise intensity differences between baseline and follow-up images, reducing the complexity of the task and mitigating generation errors. 

In contrast to other approaches, our model is conditioned on the age difference between the input and output scans, rather than on the patient's absolute age. This encourages the model to learn the relation between structural brain changes and the time interval between baseline and follow-up scans, rather than with specific age values. 
Since identical age gaps can occur at various absolute ages, this approach alleviates the need to include samples from all age groups, which is particularly advantageous when certain age ranges are underrepresented in the training data. For instance, a 5-year gap could apply both to a patient scanned at ages 60 and 65, and to another scanned at 80 and 85. While the output ages differ (65 vs. 85), the age gap remains the same (5 years).

To further improve the temporal awareness, we use a Brain-Age Estimator (BAE) to estimate the age difference between the baseline and the generated scans. During training, these predicted age differences are incorporated into the loss function, encouraging the generation of scans that accurately reflect the expected temporal interval between input and prediction.

Lastly, we introduce a \textit{Back-In-Time Regularisation} (BITR) strategy at training time, to improve the temporal awareness of the model. Based on the intuition that a temporal-aware model should be able to predict both forward and backwards in time, we randomly swap the roles of the baseline and follow-up scans, and we train the model bidirectionally to predict future or past scans, alternately. This simple regularization strategy encourages the model to learn how the brain anatomically changes going forward and backwards in time.

The method is trained and tested on the OASIS-3 dataset \cite{lamontagne2019oasis}. Moreover, we also use an external test set from the NACC dataset to evaluate the generalisation performance of TADM-3D on out-of-distribution data. As evaluation metrics, we use image-based similarity scores and brain region volumes between real and predicted follow-up MRIs. TADM-3D overcomes previous methods on both image-based and volumetric metrics, demonstrating its effectiveness on modelling the temporal brain evolution. 

Additionally, our qualitative analysis demonstrates that our method more effectively reproduces the temporal progression of brains.

This work extends our previous MICCAI 2024 conference paper \cite{Litrico24} in several ways:
\begin{itemize}
    \item The method was extended to operate on 3D MRI scans, rather than single 2D slices, allowing the model to capture richer spatial context and more complex anatomical relationships across all three dimensions.
    \item We introduced BITR, a technique aimed to improve the temporal awareness. This approach trains the model by considering not only forward progression but also backwards changes over time.
    \item The impact of the cognitive status as a conditioning variable was evaluated, allowing the model to account for clinically relevant differences. This conditioning helps capture variations in brain structure associated with different stages of cognitive decline \cite{cole2018brain}.
    \item  The comparisons were extended to include the most recent 3D methods, such as CounterSynth \cite{counterfactual_pred} and BrLP \cite{lemuel24} and we evaluate TADM-3D on an external test set to assess its generalisability.
\end{itemize}

\section{Related Works}
The study of neurodegenerative diseases through MRIs has bloomed in the last years \cite{rondinella2023enhancing}. 
Most existing approaches propose simulators that model temporal changes in brain structure. Simulators receive high-dimensional data as input, such as 3D MRIs of a given subject, and predict the changes in the brain MRIs over a specified period and under specific subject's conditions, such as neurodegenerative diseases. Most of the existing simulators are based on recent deep generative methods, including Generative Adversarial Networks (GANs) \cite{counterfactual_pred,ravi2022degenerative,Xia}, Variational Autoencoders (VAEs) \cite{vae}, Normalising Flows (NFs) \cite{wilms2022} and Diffusion Models (DMs) \cite{lemuel24,SADM}. 

Early methods \cite{ravi2022degenerative,Xia} used GANs to simulate subject-specific brain changes conditioned on the presence of a neurodegenerative disease. 
For instance, DANINet~\cite{ravi2022degenerative} employs adversarial training in conjunction with biological constraints to improve the generation process. Nevertheless, these approaches produce synthetic 2D slices, without making use of full 3D volumetric information. More recently, CounterSynth~\cite{counterfactual_pred} introduced a 3D GAN-based framework that models 3D deformations, instead of directly manipulating image pixels. 
By leveraging morphology constraints, deformations are applied to the input MRI to reflect brain changes over time. However, CounterSynth only predicts structural changes without modelling temporal evolutions. Other related approaches proposed methodologies based on VAEs \cite{vae} and NFs \cite{wilms2022}, but they produce low resolution scans and rely on morphologically constrained transformations, limiting their applicability.

More recently, the use of diffusion models has been introduced in medical image generation. DDM \cite{DDM} proposed to combine a diffusion and a deformation module to learn how to interpolate between two MRIs. This combination enables the modelling of smooth anatomical transitions by simulating plausible intermediate brain states. Similarly, DiffuseMorph \cite{DiffuseMorph} leveraged a diffusion model to predict deformation fields between two MRIs. This field is then applied to warp one image into another, effectively enabling the synthesis of interpolated images along a continuous trajectory. Both DDM and DiffuseMorph are effective in learning realistic anatomical transformations, but they are limited to interpolation tasks. As a result, their applicability in clinical settings is constrained, since they are unable to generate predictive scans.

SADM  \cite{SADM} introduced a sequence-aware diffusion model, incorporating a transformer-based architecture to better handle temporal dependencies in longitudinal data. The transformer generates a sequence of prior MRIs to extract a latent representation that encodes the subject’s disease trajectory. This representation is then used to condition the generation process of the diffusion model, allowing for the synthesis of future brain MRIs. It necessarily requires a sequence of input MRI scans, which are rarely available in a real-world context, especially at an early stage of the patient's treatment. Moreover, it does not use subject metadata, which is important to better model the disease progression. BrLP \cite{lemuel24} introduced a methodology that leverages an autoencoder, a latent diffusion model, and an auxiliary model (ControlNet) to generate individual MRI scans. However, the quality of the produced scans is highly correlated to the effectiveness of the autoencoder features. 

TADM-3D overcomes these limitations as follows: (i) unlike CounterSynth, our model learns to model the temporal progression of brain changes, making it capable of forecasting future scans; (ii) in contrast to DDM and DiffuseMorph, TADM-3D is conditioned with age differences, allowing it to predict scans at arbitrary future time points; (iii) unlike SADM, TADM-3D does not need a sequence of input MRI scans, improving its applicability in clinical context. Moreover, our model incorporates subject-specific metadata to improve the accuracy and personalisation of predictions; (iv) differently than BrLP, we train the diffusion model to directly predict MRI scans without using any auxiliary model, reducing the computational requirements and improving the prediction quality.

\section{Proposed Method}
\label{sec:proposed_method}

\subsection{Background}
DDPMs \cite{ho2020denoising} are generative models that learn how to gradually convert a Gaussian data distribution into another distribution by applying a Markov chain process. During the forward process (diffusion), Gaussian noise is progressively and incrementally added to the input data $x_0$ from the given data distribution $q(x_0)$. This occurs over a fixed number of steps, gradually transforming the sample into a latent variable distribution $q(x_t)$, as following:

\begin{equation}
    q(x_{1},x_{2},\dots,x_{T}|x_{0}) = \prod_{t=1}^{T}q(x_{t}|x_{t-1})
\end{equation}
\begin{equation}
    q(x_{t}|x_{t-1})= \mathcal{N}(x_{t};\sqrt{1-\beta_{t}}x_{t-1},\beta_{t}\textbf{I})
\end{equation}
where $t \in {1, \ldots , T}$ indicates the diffusion step, $\mathcal{N}$ is the Gaussian distribution, $\beta_{t}$ is the variance of the noise, and $\textbf{I}$ is the identity matrix. 

During the reverse process (denoising), the model is trained to learn to invert the diffusion process, by progressively turning the noise latent variable distribution $p_{\theta}(x_{t})$ into the data distribution $p_{\theta}(x_{0})$, parameterised by $\theta$. During this process, the model is trained to learn the Gaussian transformations, as follows:
\begin{equation}
    p_{\theta}(x_{0},x_{1},\dots,x_{T-1}|x_{T}) = \prod_{t=1}^{T}p_{\theta}(x_{t-1}|x_{t})
\end{equation}
\begin{equation}
    p_{\theta}(x_{t-1}|x_{t})= \mathcal{N}(x_{t-1};\mu_{0}(x_{t},t),\sigma_{0}^{2}(x_{t},t)\textbf{I})
\end{equation}
\begin{equation}
    p(x_t)= \mathcal{N}(x_T; \textbf{0}, \textbf{I})
\end{equation}

\noindent where $\sigma_{0}^{2}(x_{t},t)$ is the variance at the step $t$ and $\mu_{0}(x_{t},t)$ is the mean of the Gaussian distribution. Once trained, the model is able to denoise random noise, generating high-quality images.

\begin{figure*}[t]
        \centering
        \includegraphics[width=0.86\linewidth]{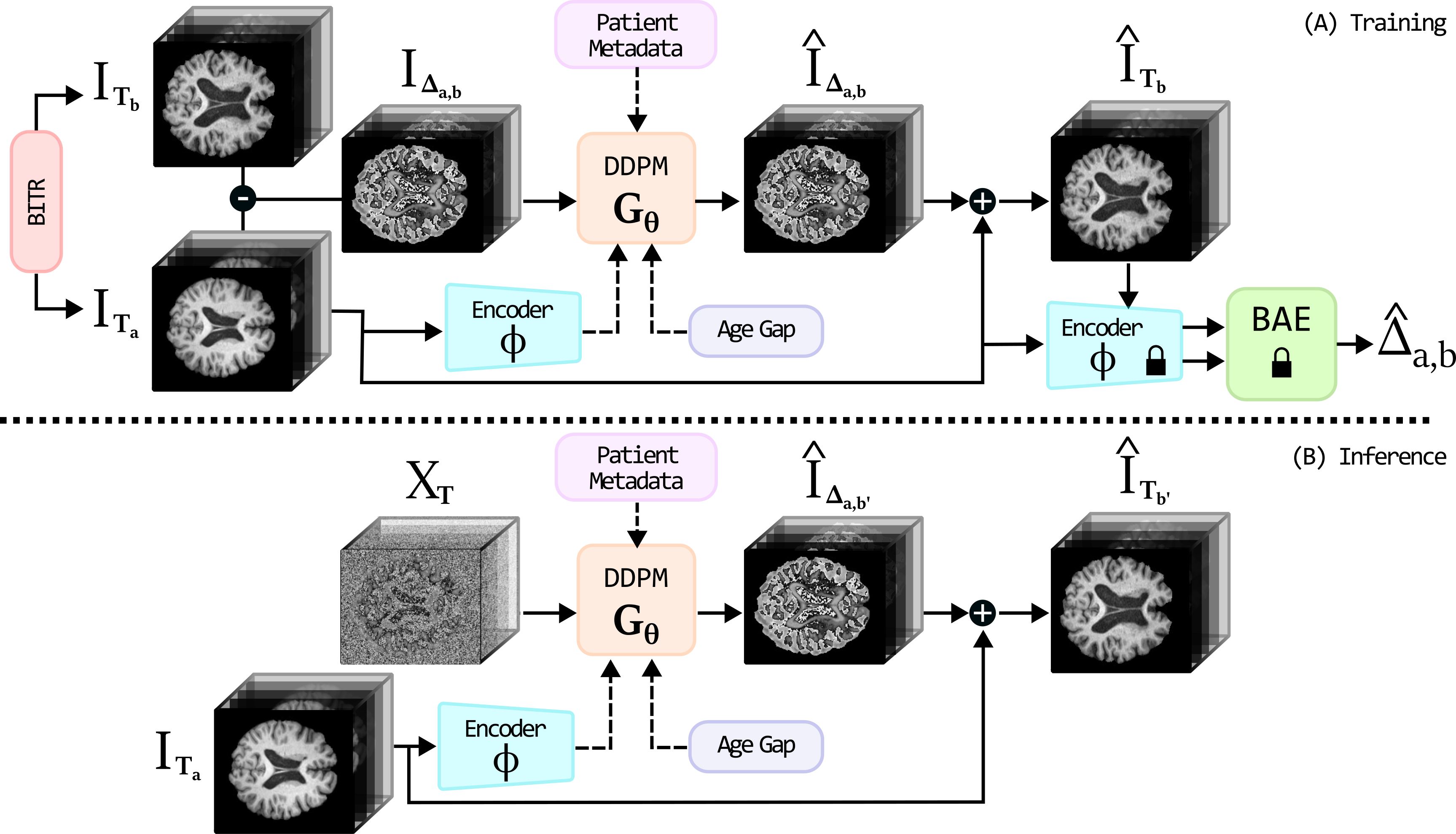}
    \caption{\textbf{TADM-3D.} (A) Given a baseline scan $I_{T_a}$ and a follow-up scan $I_{T_b}$, a residual image $I_{\Delta_{a,b}}$ is computed by subtracting the two scans. This residual is then corrupted with noise and denoised by a Denoising Diffusion Probabilistic Model (DDPM), producing the denoised residual $\widehat{I}_{\Delta{a,b}}$. The denoised residual image and the scan $I_{T_a}$ are added together to estimate the scan $\widehat{I}_{T_b}$ at time $T_{b}$ (see \cref{subsec:tadm}). The baseline scan is encoded with $\Phi$ to produce a latent representation $z_a$, which, along with patient metadata, is used to condition the DDPM (see \cref{sec:conditioning}). The estimated follow-up scan $\widehat{I}_{T_b}$ is also encoded to extract $z_b$. Representations $z_a$ and $z_b$ are then fed into a BAE to predict the time interval $\widehat{\Delta}_{a,b}$ between the scans (see \cref{sec:tge}). To promote time awareness, the model is also trained to predict past scans with a probability $p=0.5$ by swapping the roles of baseline and follow-up scans (see \cref{sec:regularisation}). The \textsc{padlock} indicates frozen parameters. Dashed lines indicate model conditioning. (B) At inference time, given the baseline scan $I_{T_a}$ and random noise $X_T$, the DDPM predicts the residual $\mathbf{\widehat{I}}_{\Delta_{a,b'}}$ that summed to the baseline $I_{T_a}$ produces the predicted follow-up $\mathbf{\widehat{I}}_{T_{b'}}$.}
    \label{fig:pipeline}
\end{figure*}

\subsection{Temporally-Aware Diffusion Model}
\label{subsec:tadm}
Our proposed training pipeline, depicted in \Cref{fig:pipeline}, comprises three main components: (i) a Denoising Diffusion Probabilistic Model (DDPM), responsible for modelling the brain progression and generating high quality scans; (ii) an Encoder, which extracts latent representation from MRI scans; and (iii) a Brain-Age Estimator (BAE), which is leveraged to encourage temporal consistency by estimating the time interval between scans.

During training, we use MRI pairs, \textit{i.e.} $I_{T_a}$ and $I_{T_b}$, corresponding to scans acquired from the same subject at two distinct timepoints $T_a$ and $T_b$, respectively. From these images, we compute a residual image $I_{\Delta_{a,b}} = I_{T_b} - I_{T_a}$, which captures the voxel-wise intensity changes occurring over the time interval $\Delta_{a,b} = T_b - T_a$. This residual is used to train the DDPM at predicting the residual $\widehat{I}_{\Delta_{a,b}}$. To generate the output follow-up scans $\widehat{I}_{T_b}$, we add the predicted residual to the baseline scan, such that $\widehat{I}_{T_b} = I_{T_a} + \widehat{I}_{\Delta{a,b}}$. Additionally, we leverage the BAE model to estimate the time interval $\widehat{\Delta}_{a,b}$ between $I_{T_a}$ and the estimated $\widehat{I}_{T_b}$. The estimated time interval is then integrated in the loss function to regularise the DDPM generation process. At inference, we utilize the scan $I_{T_a}$ acquired at time $T_a$ along with the target time interval $\Delta_{a,b'}$ to produce a future scan $I_{T_{b'}}$ at time $T_{b'}$. To achieve patient individualization during the generation process, the DDPM is conditioned on a latent embedding obtained by applying an encoder $\Phi$ to the baseline scan $I_{T_a}$, along with auxiliary patient-specific metadata (\textit{e.g.}, cognitive status and age).

\subsection{Conditioning Strategies}
\label{sec:conditioning}
To generate residual images, we condition the DDPM using: (i) baseline latent representation $\Phi(I_{T_a})$ extracted by the encoder $\Phi$ on the baseline scan $I_{T_a}$; (ii) the time interval $\Delta_{a,b}$ between $T_a$ and $T_b$; (iii) patient's specific metadata.

\mypar{Baseline Scan Encoding} Our goal is to predict patient-specific brain trajectories. To achieve this, the DDPM is conditioned with the encoded representation $z_a$ derived from the baseline scan $I_{T_a}$, using an encoder $\Phi$. This latent features capture the patient-specific anatomy, enabling the generation of personalised outputs.

\mypar{Age Difference}
Using age directly to model progression fails to explicitly capture the temporal evolution of structural changes in brain MRIs, and it demands age-balanced datasets, which are often unavailable in real-world settings. To address this issue, we propose conditioning the model on the age difference between scans, denoted as $\Delta_{a,b}$. Since the same age gaps can arise from scan pairs acquired at different ages, this strategy reduces the need of uniformly sampling all age groups during training. This is especially beneficial when certain age ranges are underrepresented in the dataset. In our approach, we encode $\Delta_{a,b}$ using positional encoding \cite{transformer} before integrating it into the model.
 
\mypar{Patient-Specific Metadata} We also use both the age ($A$) of the patient at time $T_a$, and cognitive status ($D$) to further condition our model. Baseline age provides essential context to the model, as neurodegenerative diseases progress differently depending on the age of the patient. Note that, unlike previous approaches that condition using the age of the patient w.r.t. the follow-up scan, we use the age at baseline to accurately ground the starting point of the disease trajectory. 

\subsection{Estimating Brain Age for Improving the Temporal Awareness} \label{sec:tge}
To guide the model in generating scans that accurately reflect the expected age difference between the input and predicted scans, we incorporate a Brain Age Estimator (BAE) \cite{jonsson2019brain} into our training pipeline. During DDPM training, the BAE evaluates whether the generated follow-up scan exhibits brain changes consistent with the expected time interval from the baseline. By including the predicted age difference as part of the loss function, this mechanism encourages the diffusion model to generate outputs that not only appear realistic but also accurately capture the desired temporal progression. This approach provides feedback to the model, thereby enhancing the temporal consistency of the predictions. The BAE model is pre-trained on the training set, and its parameters remain fixed during the DDPM training. 

Given a baseline scan $I_{T_a}$ and a generated follow-up scan $\widehat{I}_{T_b}$, we compute the estimated age difference as $\widehat{\Delta}_{a,b} = \Psi(\Phi(\widehat{I}_{T_b})) - \Psi(\Phi(I_{T_a}))$, where $\Phi$ is the encoder and $\Psi$ is the BAE model. When the DDPM prediction $\widehat{I}_{T_b}$ accurately matches the ground-truth image $I_{T_b}$, the estimated age difference $\widehat{\Delta}_{a,b}$ aligns with the actual age gap $\Delta_{a,b}$, ensuring that the generated output reflects the correct temporal progression. Any differences between the predicted and true age gaps are backpropagated during training, serving as a regularization. This feedback loop iteratively refines the generation process, enhancing the model's ability to produce scans that accurately reflect the temporal progression.

\subsection{Back-In-Time Regularisation} \label{sec:regularisation}
TADM-3D is trained using pairs of MRI scans $I_{T_a}$ and $I_{T_b}$ acquired from the same patient at two distinct time points. Here $I_{T_a}$ indicates the baseline scan acquired at age $A$, and $I_{T_b}$ is the follow-up scan acquired after a time interval $\Delta_{a,b}$. In this way, the diffusion model learns to predict the brain structural changes that occur due to the combined effect of ageing and disease progression over the time interval. However, a temporal-aware model should also be able to predict past scans, \textit{e.g.} generating a baseline scan $I_{T_a}$ from a follow-up scan $I_{T_b}$.

To this aim, we propose to include a \textit{Back-In-Time Regularisation} (BITR) strategy. Specifically, at each training step, we randomly swap the roles of the two scans $I_{T_a}$ and $I_{T_b}$ with a probability $p=0.5$.
Consequently, the diffusion model is trained to predict the follow-up from the baseline scan for approximately half of the training iterations and the baseline from the follow-up in the other half of the training steps. Note that this swapping is exclusively applied in the training phase. Although simple, this strategy encourages the model to learn a better relationship between anatomical changes in the brain and time interval, leading to an improved temporal awareness.

\subsection{Overall Framework}

\begin{algorithm}[t!]
\footnotesize
\begin{algorithmic}
\State \textbf{Input:} Pairs of MRI scans $(I_{T_a}, I_{T_b})$, time interval $\Delta_{a,b}$, patient metadata (age at baseline $A$, cognitive status $D$).
\State \textbf{Output:} Trained DDPM model $G_\theta$ for generating future MRI scans

\State \textbf{Training:}
\For{each training step}
    \State Sample a pair $(I_{T_a}, I_{T_b})$ from the training set
    \State Compute residual image $I_{\Delta_{a,b}} = I_{T_b} - I_{T_a}$
    \State Extract latent representation 
    \Statex \qquad $z_a = \Phi(I_{T_a})$ using the encoder $\Phi$
    \State Sample $p\sim \mathcal{B}(0.5)$
    \If{$p=1$}
        \State Flip sign in $\Delta_{a,b}$, $A$, and $D$
    \EndIf
    
    \State Sample noise $\epsilon \sim \mathcal{N}(0,1)$
    \State Compute noisy residual $\mathbf{\bar{I}}_{\Delta_{a,b}}$ at diffusion step $t$
    \State Predict noise using DDPM: 
    \Statex \qquad $\hat{\epsilon} = G_\theta(\mathbf{\bar{I}}_{\Delta_{a,b}}, t; z_a, \Delta_{a,b}, A, D)$
    \State Compute DDPM loss: $\mathcal{L}^{DML} = ||\hat{\epsilon} - \epsilon||_2^2$
    \State Generate predicted residual $\widehat{I}_{\Delta_{a,b}}$ from the DDPM
    \State Compute predicted follow-up scan $\widehat{I}_{T_b} = I_{T_a} + \widehat{I}_{\Delta_{a,b}}$
    \State Extract latent representation $z_b = \Phi(\widehat{I}_{T_b})$
    \State Predict age difference using BAE: $\widehat{\Delta}_{a,b} = \Psi(z_b) - \Psi(z_a)$
    \State Compute BAE loss: $\mathcal{L}^{BAE} = (\widehat{\Delta}_{a,b} - \Delta_{a,b})^2$
    \State Compute total loss: $\mathcal{L}^{Tot} = \mathcal{L}^{DML} + \mathcal{L}^{BAE}$
    \State Update DDPM parameters $\theta$ using gradients of $\mathcal{L}^{Tot}$
\EndFor

\end{algorithmic}
\caption{Pseudocode of TADM-3D's training process.}
\label{alg:tadm}
\end{algorithm}

\mypar{Training} A complete overview of the training process is described in \Cref{alg:tadm}. 
During the diffusion process, the DDPM learns to estimate the noise $\epsilon$ incorporated into $I_{\Delta_{a,b}}$. This is achieved by minimising the following loss function:

\begin{equation}
    \label{eq:dml}
    \mathcal{L}^{DML} = \mathbb{E}_{\epsilon \sim \mathcal{N}(0,1), \mathbf{\bar{I}}_{\Delta_{a,b}}, t} \left[||G_\theta( \mathbf{\bar{I}}_{\Delta_{a,b}}, t; z_a, \Delta_{a,b}, A, D) - \epsilon ||^2_2 \right],
\end{equation}

\noindent where, $G_\theta$ represents the DDPM with parameters $\theta$, and $t$ denotes the diffusion timestep. Furthermore, as discussed in \Cref{sec:tge}, the prediction of BAE is incorporated as an extra component in the loss function. Specifically, we define the loss on the expected brain age difference as follows: 
\begin{equation}
    \label{eq:tge}
    \mathcal{L}^{BAE} = (\widehat{\Delta}_{a,b} - \Delta_{a,b})^2.
\end{equation}

\noindent Finally, the overall loss is obtained by combining \cref{eq:dml,eq:tge}:
\begin{equation}
    \mathcal{L}^{Tot} = \mathcal{L}^{DML} + \mathcal{L}^{BAE}
\end{equation}

\mypar{Inference} Given a baseline MRI $\mathbf{I}_{T_a}$, the model generates a future MRI $\mathbf{\widehat{I}}_{T_{b'}}$ at any desired time interval $\Delta_{a,b'}$ after the baseline. The generation process starts with a random Gaussian noise input $X_T$, which is iteratively refined by the network $G_\theta( X_t, t; z_a, \Delta_{a,b'}, A, D)$. The generated residual image $\mathbf{\hat{I}}_{\Delta{a,b'}}$ is finally combined with the baseline scan to yield the follow-up prediction.

\section{Experimental Results}
\label{sec:experimental_results}

\subsection{Datasets} We train TADM-3D on 2,535 T1-weighted (T1w) brain MRI scans from 634 subjects from the  OASIS-3 dataset~\cite{lamontagne2019oasis}. Scans span a longitudinal interval of approximately $15$ years, capturing a broad spectrum of aging-related changes. The dataset includes participants aged between $42$ and $95$ years classified as cognitively normal (CN), mild cognitive impairment (MCI), and Alzheimer's disease (AD). We evaluate TADM-3D on both internal and external test sets, to also asses generalisation performance with out-of-distribution data. For the external dataset, we used data from the NACC dataset \cite{nacc}, including 2,257 T1w MRIs from 962 subjects. Scan have a maximum interval between the initial and follow-up MRI of $13$ years, with an average of $3.8$. About $75\%$ of the patients are classified as CN at the last visit, while the remaining $25\%$ are MCI or AD. Data from the external dataset is used solely for evaluation.

\subsection{Implementation Details} All MRI volumes were linearly registered using the MNI152 template to normalise spatial orientation and scale, and skull stripping using the FSL library~\cite{jenkinson2012fsl}. The dataset is divided into a training set (70\%), a validation set (10\%) and a test set (20\%). For both the training of diffusion and BAE models, we use the AdamW optimiser with a learning rate of $0.0001$ and a weight decay of $0.001$, a batch size of $16$, and a cosine-based learning rate scheduler. Following \cite{li2022srdiff}, we employ the U-Net architecture as backbone of the diffusion model $G_\theta$. For the encoder $\Phi$, we use a 3D UNet-based architecture trained together with the diffusion model in an end-to-end manner. Results for all state-of-the-art methods were obtained using their publicly available implementations.

\begin{table*}[t]
    \centering
    \setlength{\tabcolsep}{5pt}
    \def\arraystretch{1.5}
    \resizebox{\linewidth}{!}{
    \begin{tabular}{l|cc|ccccc}
    \toprule 
          &&& \multicolumn{5}{|c}{\textbf{Region Volumes Error} (\%) $\downarrow$} \\
        Method  & MSE $\downarrow$ & SSIM $\uparrow$ & Hippocampus & Amygdala & Lat. Ventricle & Thalamus & CSF   \\
        \midrule
        DaniNet \cite{ravi2022degenerative}  & 0.016 $\pm$ 0.007 & 0.623 $\pm$ 0.162 & 0.030 $\pm$ 0.030 & 0.018 $\pm$ 0.017 & 0.257 $\pm$ 0.222 & 0.038 $\pm$ 0.030 & 1.081 $\pm$ 0.814 \\
        CounterSynth \cite{counterfactual_pred} & 0.008 $\pm$ 0.004 & 0.861 $\pm$ 0.052 & 0.030 $\pm$ 0.018 & 0.016 $\pm$ 0.010 & 0.273 $\pm$ 0.311 & 0.041 $\pm$ 0.035 & 0.881 $\pm$0.672 \\
        BrLP \cite{lemuel24} & 0.005 $\pm$ 0.002 & 0.887 $\pm$ 0.017 & 0.028 $\pm$ 0.018 & 0.017 $\pm$ 0.009 & 0.264 $\pm$ 0.271 & 0.039 $\pm$ 0.021 & 0.882 $\pm$ 0.645 \\
        \textbf{TADM-3D} & \textbf{0.004 $\pm$ 0.001} & \textbf{0.902 $\pm$ 0.014} & \textbf{0.017 $\pm$ 0.019} & \textbf{0.015 $\pm$ 0.014} & \textbf{0.228 $\pm$ 0.187} & \textbf{0.027 $\pm$ 0.015} & \textbf{0.642 $\pm$ 0.412} \\
        \bottomrule
    \end{tabular}
    }
    \caption{Results on the internal test set of OASIS-3. Performance are evaluated in terms of image-based and region volumes errors wth respect to other methods. The MAE in region volumes is expressed as a percentage of total brain volume. }
    \label{tab:oasis_results}
\end{table*}

\begin{table*}[t]
    \centering
    \setlength{\tabcolsep}{5pt}
    \def\arraystretch{1.5}
    \resizebox{\linewidth}{!}{
    \begin{tabular}{l|cc|ccccc}
    \toprule 
          &&& \multicolumn{5}{|c}{\textbf{Region Volumes Error} (\%) $\downarrow$} \\
        Method  & MSE $\downarrow$ & SSIM $\uparrow$ & Hippocampus & Amygdala & Lat. Ventricle & Thalamus & CSF   \\
        \midrule
        DaniNet \cite{ravi2022degenerative}  & 0.017 $\pm$ 0.007 & 0.611 $\pm$ 0.181 & 0.032 $\pm$ 0.031 & 0.018 $\pm$ 0.016 & 0.232 $\pm$ 0.210 & 0.039 $\pm$ 0.032 & 1.154 $\pm$ 0.871 \\
        CounterSynth \cite{counterfactual_pred} &  0.011 $\pm$ 0.003 & 0.813 $\pm$ 0.042 & 0.030 $\pm$ 0.020 & 0.014 $\pm$ 0.010 & 0.283 $\pm$ 0.314 & 0.111 $\pm$ 0.034 & 1.173 $\pm$ 0.731 \\
        BrLP \cite{lemuel24} & 0.005 $\pm$ 0.002 & \textbf{0.909 $\pm$ 0.023} & 0.024 $\pm$ 0.023 & 0.014 $\pm$ 0.013 & \textbf{0.213 $\pm$ 0.350} & 0.030 $\pm$ 0.024 & 1.044 $\pm$ 0.788 \\
        \textbf{TADM-3D} & \textbf{0.004 $\pm$ 0.002} & 0.902 $\pm$ 0.017 & \textbf{0.020 $\pm$ 0.020} & \textbf{0.013 $\pm$ 0.011} & 0.235 $\pm$ 0.200 & \textbf{0.029 $\pm$ 0.021} & \textbf{0.833 $\pm$ 0.543} \\
        \bottomrule
    \end{tabular}
    }
    \caption{Results on the external test set of NACC evaluating TADM-3D's generalisation performance in terms of image-based and region volumes errors in comparison to other methods. The MAE in region volumes is expressed as a percentage of total brain volume. }
    \label{tab:nacc_results}
\end{table*}

\subsection{Evaluation Metrics} To evaluate the performance of our method, we employed both image-based similarity metrics and region-specific volumetric analyses in anatomically relevant brain areas. For the image-based evaluation, we use the \textit{Structural Similarity Index Measure} (SSIM) and \textit{Mean Squared Error} (MSE), measuring the similarity between the generated and ground-truth MRI scans in terms of structural fidelity and pixel-wise reconstruction accuracy, respectively. On the other side, volumetric metrics in AD-related regions (lateral ventricles, cerebrospinal fluid as CSF, hippocampus, amygdala, and thalamus) are computed to evaluate the modelling of disease progression in TADM-3D. We use \textit{SynthSeg} 2.0 \cite{billot_synthseg_2023} to segment the brain and compute the region volumes, which are expressed as percentages of the entire brain to account for personal variations.

The accuracy of predicted regional volumes is then assessed by computing the Mean Absolute Error (MAE) with the ground-truth volumes.

\begin{figure*}[t]
    \centering
    \includegraphics[width=1\linewidth]{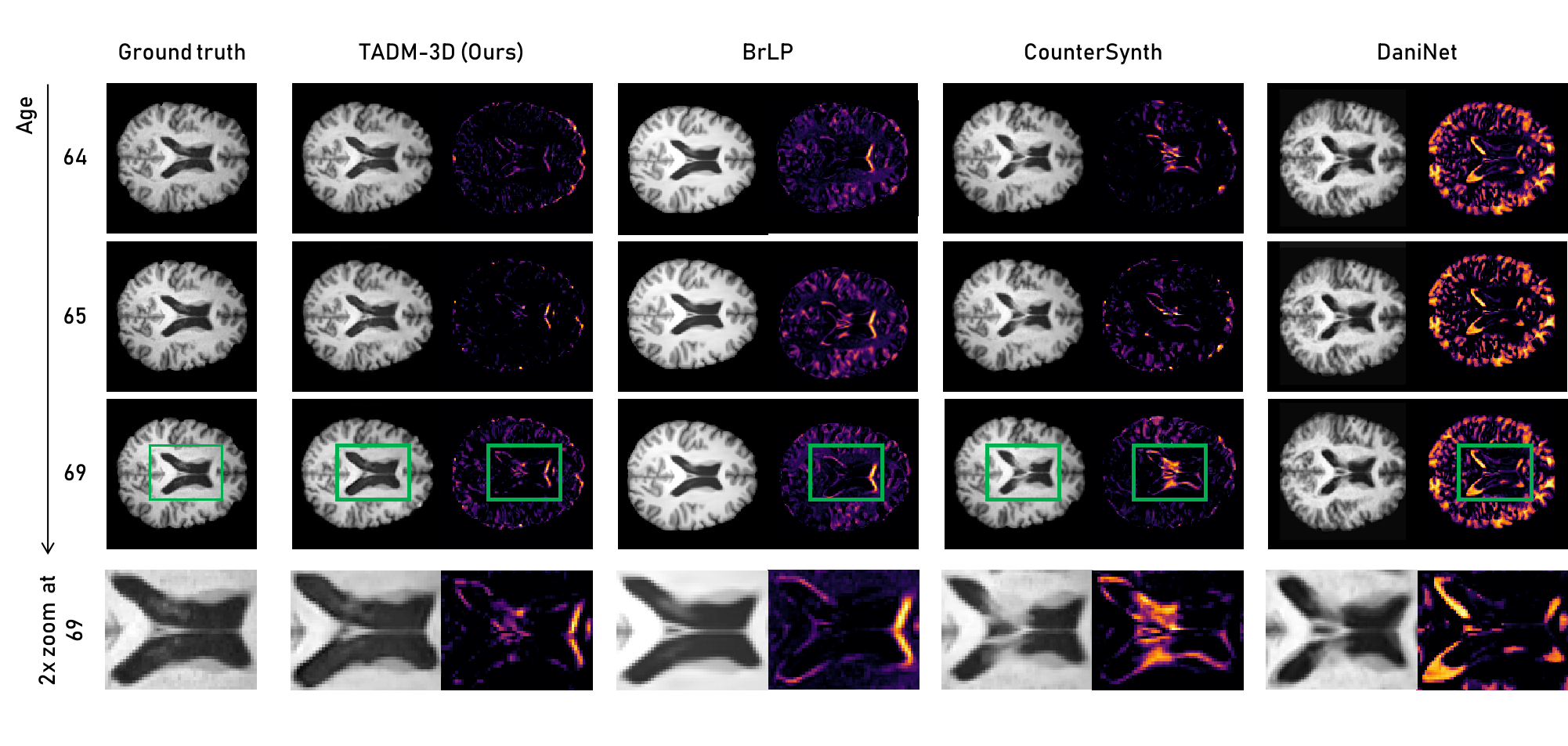}
    \caption{Temporal progression on a 62-year-old subject with AD from the internal test set, generated by our approach against BrLP\cite{lemuel24}, CounterSynth\cite{counterfactual_pred} and DaniNet\cite{ravi2022degenerative}. We show predicted MRIs on the centre slice on the left and the corresponding heatmap of prediction error on the right.}
    \label{fig:qualitative_results}
\end{figure*}

\subsection{Comparative Analysis} \Cref{tab:oasis_results} shows the quantitative results on the internal test set obtained by TADM-3D compared to other state-of-the-art 3D approaches \cite{counterfactual_pred,lemuel24,ravi2022degenerative,SADM}. Overall, TADM-3D outperforms them across the board, reducing the MSE and increasing the SSIM by $+0.001$ and $+0.15$, respectively. Results on the region's volumes show that TADM-3D achieves the lowest error in different brain regions. In particular, we reduce the error on Hippocampus, Amygdala and Ventricles by approximately $40\%$, $12\%$, and $13\%$, respectively. On Thalamus and CSF, TADM-3D reduces the error by $30\%$ and $27\%$, respectively. These results show that TADM-3D generates accurate follow-up images with respect to existing approaches, achieving better performance in both image-based and volumetric metrics. On the external test set (see \Cref{tab:nacc_results}), TADM-3D obtains similar performance w.r.t. the internal evaluation, showing robustness to generalisation challenges. TADM-3D achieves the best performance on the MSE and on volumetric errors of 4 out of 5 analysed brain regions, while BrLP unexpectedly improves its results w.r.t. the internal test set on SSIM and on volumetric error of the Lateral Ventricle, obtaining the lowest error. 

Furthermore, \Cref{fig:qualitative_results} presents a sample of qualitative results illustrating the prediction of MRI scans at various ages for a 62-year-old patient diagnosed with AD.

Clearly, TADM-3D offers a better approximation of the temporal evolution of the brain compared to all other methods. Specifically, our predictions show a notable accuracy in modelling the ventricular expansion over time, maintaining a lower error at every age difference. In comparison to BrLP \cite{lemuel24} and DaniNet \cite{ravi2022degenerative}, our model reaches superior performance, especially on the ventricular region. CounterSynth \cite{counterfactual_pred} seems to produce similar results, but TADM-3D exhibits more balanced and consistently lower errors.

\begin{table*}[t]
    \centering
    \setlength{\tabcolsep}{5pt}
    \def\arraystretch{1.5}
    \resizebox{\linewidth}{!}{
    \begin{tabular}{l|cc|ccccc}
        \toprule
        &&& \multicolumn{5}{|c}{\textbf{Region Volumes Error} (\%) $\downarrow$} \\
        Method  & MSE $\downarrow$ & SSIM $\uparrow$ & Hippocampus & Amygdala & Lat. Ventricle & Thalamus & CSF   \\
        \midrule
        TADM-3D w/o patient metadata & 0.006 $\pm$ 0.001 & 0.899 $\pm$ 0.017 & 0.019 $\pm$ 0.011 & \textbf{0.015 $\pm$ 0.018} & 0.235 $\pm$ 0.201 & 0.030 $\pm$ 0.019 & 0.648 $\pm$ 0.557\\
        TADM-3D w/o age gaps cond. & 0.011 $\pm$ 0.006 & 0.812 $\pm$ 0.016 & 0.027 $\pm$ 0.013 & 0.017 $\pm$ 0.021 & 0.251 $\pm$ 0.231 & 0.037 $\pm$ 0.029 & 0.747 $\pm$ 0.555 \\
        TADM-3D w/o BAE  & 0.009 $\pm$ 0.005 & 0.872 $\pm$ 0.024 & 0.025 $\pm$ 0.014 & 0.016 $\pm$ 0.025 & 0.249 $\pm$ 0.212 & 0.036 $\pm$ 0.026 & 0.691 $\pm$ 0.512 \\
        TADM-3D w/o BITR  & 0.008 $\pm$ 0.003 & 874 $\pm$ 0.021 & 0.023 $\pm$ 0.016 & 0.016 $\pm$ 0.022 & 0.241 $\pm$ 0.200 & 0.031 $\pm$ 0.021 & 0.667 $\pm$ 0.491 \\
        TADM-2.5D & 0.014 $\pm$ 0.006 & 0.779 $\pm$ 0.089 & 0.030 $\pm$ 0.027 & 0.016 $\pm$ 0.015 & 0.299 $\pm$ 0.277 & 0.060 $\pm$ 0.041 & 0.899 $\pm$ 0.643\\
        \midrule
        \textbf{TADM-3D} & \textbf{0.004 $\pm$ 0.001} & \textbf{0.902 $\pm$ 0.014} & \textbf{0.017 $\pm$ 0.019} & \textbf{0.015 $\pm$ 0.014} & \textbf{0.228 $\pm$ 0.187} & \textbf{0.027 $\pm$ 0.015} & \textbf{0.642 $\pm$ 0.412} \\
        \bottomrule
    \end{tabular}
    }
    \caption{Ablation studies: showing the results of the absence of the different components of TADM-3D. Finally, we also show the impact of using a 2D model to individually generate scans rather than a native 3D architecture.}
    \label{tab:add_analysis}
\end{table*}

\subsection{Ablation Studies}
\label{sec:ablation} 
In this section, we present an ablation study to assess the contribution of the key components of TADM-3D. Specifically, we evaluate the effect of removing conditioning on age differences, patient-specific conditioning variables, and excluding the BAE module and BITR from training. Furthermore, we compare the performance of our 3D generative architecture against a 2D alternative, where slices are generated independently to reconstruct full 3D volumes (named TADM-2.5D). 

\Cref{tab:add_analysis} (first row) presents the performance of TADM-3D without using patient metadata. Results show that, removing the conditioning with such metadata, we observe a minimal performance loss. In the second row, we observe that removing the conditioning of the model on age gaps highly drops performance, demonstrating the effectiveness of our strategy over the current state-of-the-art. Another case of performance drop is when BAE is not used (third row), indicating its usefulness in supporting the generation process. In the fourth row, we show the impact of removing the BITR with a significant drop of performance, demonstrating its contribution in improving the training. Finally, in the last row, we assess the performance using a 2D diffusion model (as our previous work \cite{Litrico24}) rather than a 3D one. To reconstruct full 3D volumes, we train a 2D model to generate each slice independently, using the slice index as a conditioning variable to preserve spatial coherence across the volume. Results show a drastically reduction in performance when using the 2D model, demonstrating that a 3D model captures better structural information during training.

\begin{table*}[t]
    \caption{Evaluating the impact of incorrect conditioning on cognitive status in TADM-3D predictions.}
    \centering
    \setlength{\tabcolsep}{5pt}
    \def\arraystretch{1.5}
    \resizebox{\linewidth}{!}{
    \begin{tabular}{l|cc|ccccc}
        \toprule
        &&& \multicolumn{5}{|c}{\textbf{Region Volumes Error} (\%) $\downarrow$}\\
        Method  & MSE $\downarrow$ & SSIM $\uparrow$ & Hippocampus & Amygdala & Lat. Ventricle & Thalamus & CSF   \\
        \midrule
        TADM-3D w/ wrong cond. & 0.006 $\pm$ 0.005 & 0.895 $\pm$ 0.012 & 0.030 $\pm$ 0.013 & 0.017 $\pm$ 0.021 & 0.234 $\pm$ 0.187 & 0.030 $\pm$ 0.026 & 0.679 $\pm$ 0.417\\
        \textbf{TADM-3D} & \textbf{0.004 $\pm$ 0.001} & \textbf{0.902 $\pm$ 0.014} & \textbf{0.017 $\pm$ 0.019} & \textbf{0.015 $\pm$ 0.014} & \textbf{0.228 $\pm$ 0.187} & \textbf{0.027 $\pm$ 0.015} & \textbf{0.642 $\pm$ 0.412} \\
        \bottomrule
    \end{tabular}
    }
    \label{tab:wrong_cond}
\end{table*}

\subsection{Is TADM-3D Modelling the Disease Progression?}
Following the protocol proposed in \cite{puglisi2025brain}, we perform an experiment to investigate the potential prediction bias of TADM-3D toward modelling healthy ageing trajectories. Specifically, we aim to assess whether the model’s outputs are appropriately generated with respect to the patient's cognitive status. The goal is to assess whether TADM-3D can effectively learn to generate normal and pathological trajectories. To do so, we use our trained model with scans obtained from patients with AD, but we deliberately condition the model, specifying that the patient is Cognitively Normal (CN). We then generate predicted future scans under both the incorrect (CN) and correct (AD) cognitive conditions, and compare the resulting volumetric predictions. This setup allows us to isolate the influence of cognitive status on the generative behaviour of the model. Results in \cref{tab:wrong_cond} show that prediction errors generally increase when incorrect cognitive conditioning is applied, particularly in the hippocampal region, a structure notably affected in AD. This supports the hypothesis that the model is not biased toward healthy ageing and is capable of capturing distinct anatomical progressions associated with neurodegeneration. 

\section{Discussions}
\subsection{Limitations}
\begin{figure*}[t]
    \centering
    \includegraphics[width=1\linewidth]{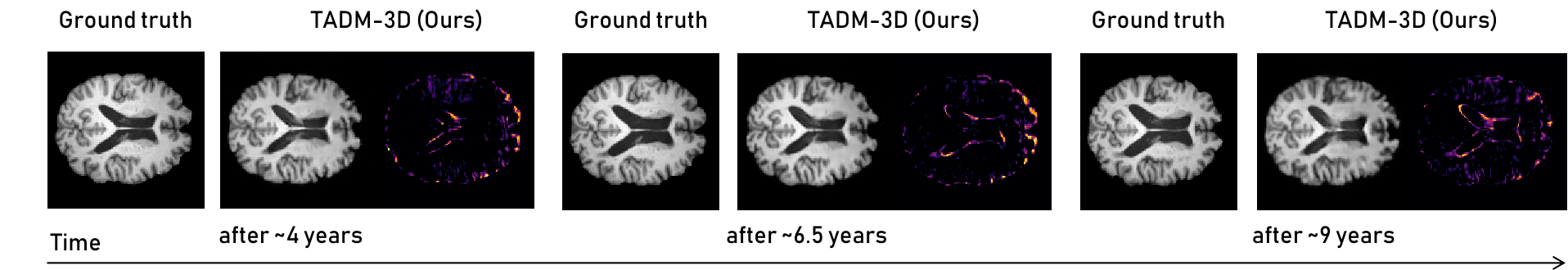}
    \caption{Example of limitation of TADM-3D in predicting long-time trajectory for a 65-year-old patient. The brain's evolution in predictions after 6.5 and 9 years is inconsistent, showing limitations in capturing long-time dependencies. }
    \label{fig:limitations}
\end{figure*}

Despite the promising results achieved by TADM-3D, some limitations still warrant further investigation. Specifically, \Cref{fig:limitations} illustrates an example of the challenges faced by TADM-3D in predicting long-term trajectories for a 65-year-old patient. The model tends to lose accuracy when predicting brain evolution over long time intervals, exhibiting increased errors, particularly for intervals spanning several years between scans. This highlights a difficulty in effectively capturing long-term temporal dependencies and predicting distant future anatomical changes. Another limitation is that, although TADM-3D is conditioned on age differences and some patient metadata, it does not yet incorporate other important clinical factors such as genetic information, comorbidities, or medication effects, that could significantly influence disease trajectories.

\subsection{Applications}
Disease progression modelling plays a crucial role in many aspects of clinical practice and holds significant promise for improving patient care. By simulating the anatomical changes that occur over time, TADM-3D enables earlier diagnosis and more accurate prognostic assessments, which are essential for timely and targeted interventions. Predicting personalised progression trajectories allows for individualized treatment planning, thereby optimising care strategies. In the context of clinical trials, our model can generate virtual follow-up scans or synthetic control arms, which can enhance trial design and reduce patient burden. TADM-3D also helps overcome limitations caused by missing or irregularly acquired longitudinal scans by effectively filling gaps and supporting robust longitudinal analyses. From a research perspective, modelling disease trajectories provides insights into the underlying pathophysiological mechanisms and temporal dynamics, contributing to novel biomarker discovery. Moreover, our model supports educational efforts by delivering visual representations of expected progression, improving communication among clinicians, patients, and their families. Finally, TADM-3D may help address data scarcity and biases in medical research by synthesising data for underrepresented populations and costly imaging modalities, including uncommon and expensive scans such as PET and CT, where generating synthetic samples remains particularly challenging.

In summary, these applications highlight the importance of our model for precision medicine, enabling earlier, more personalised, and more effective care for progressive diseases such as Alzheimer's.

\subsection{Future Works}
Building upon our pipeline, multiple directions could be followed to further advance TADM-3D. One direction is to enhance the model's ability to capture longer-term temporal dependencies. Another one involves integrating multiple data modalities beyond structural MRI, such as other medical scans, genomic information and clinical records. This multi-modal integration could provide more comprehensive and personalised information on the neurodegenerative progression. Additionally, improving the model’s robustness and adaptability to diverse clinical and imaging settings could be beneficial. Indeed, approaches based on federated learning frameworks could facilitate model deployment across institutions without requiring data sharing. Finally, extending the framework to other organs would further broaden its clinical impact.

\section{Conclusion}
\label{sec:conclusion}
In this work, we introduced TADM-3D, a diffusion-based approach for 3D brain progression modelling that directly predicts the intensity difference between baseline and follow-up MRI scans, effectively capturing structural changes over time. Our method addresses several key limitations of existing techniques by conditioning predictions on age differences rather than on target ages, allowing for more accurate temporal modelling without requiring age-balanced datasets. Furthermore, we proposed a Back-In-Time Regularization strategy, enhancing the model's temporal awareness by training the model to predict both forward and backwards in time. 

We extensively evaluated TADM-3D on the OASIS-3 dataset, achieving superior results in both similarity metrics and region volume estimation. Moreover, to assess the generalisability of our method, we also tested TADM-3D on an external test set from the NACC dataset, achieving comparable results w.r.t. the internal evaluation. Qualitative analyses further highlighted the model’s capacity to better capture the progression of brain structures over time, particularly in regions highly susceptible to neurodegenerative diseases.

\section*{Acknowledgment}

The NACC database is funded by NIA/NIH Grant U24 AG072122. NACC data are contributed by the NIA-funded ADRCs: P30 AG062429 (PI James Brewer, MD, PhD), P30 AG066468 (PI Oscar Lopez, MD), P30 AG062421 (PI Bradley Hyman, MD, PhD), P30 AG066509 (PI Thomas Grabowski, MD), P30 AG066514 (PI Mary Sano, PhD), P30 AG066530 (PI Helena Chui, MD), P30 AG066507 (PI Marilyn Albert, PhD), P30 AG066444 (PI David Holtzman, MD), P30 AG066518 (PI Lisa Silbert, MD, MCR), P30 AG066512 (PI Thomas Wisniewski, MD), P30 AG066462 (PI Scott Small, MD), P30 AG072979 (PI David Wolk, MD), P30 AG072972 (PI Charles DeCarli, MD), P30 AG072976 (PI Andrew Saykin, PsyD), P30 AG072975 (PI Julie A. Schneider, MD, MS), P30 AG072978 (PI Ann McKee, MD), P30 AG072977 (PI Robert Vassar, PhD), P30 AG066519 (PI Frank LaFerla, PhD), P30 AG062677 (PI Ronald Petersen, MD, PhD), P30 AG079280 (PI Jessica Langbaum, PhD), P30 AG062422 (PI Gil Rabinovici, MD), P30 AG066511 (PI Allan Levey, MD, PhD), P30 AG072946 (PI Linda Van Eldik, PhD), P30 AG062715 (PI Sanjay Asthana, MD, FRCP), P30 AG072973 (PI Russell Swerdlow, MD), P30 AG066506 (PI Glenn Smith, PhD, ABPP), P30 AG066508 (PI Stephen Strittmatter, MD, PhD), P30 AG066515 (PI Victor Henderson, MD, MS), P30 AG072947 (PI Suzanne Craft, PhD), P30 AG072931 (PI Henry Paulson, MD, PhD), P30 AG066546 (PI Sudha Seshadri, MD), P30 AG086401 (PI Erik Roberson, MD, PhD), P30 AG086404 (PI Gary Rosenberg, MD), P20 AG068082 (PI Angela Jefferson, PhD), P30 AG072958 (PI Heather Whitson, MD), P30 AG072959 (PI James Leverenz, MD).


\bibliographystyle{elsarticle-num} 
\bibliography{main.bib}






\end{document}